\documentclass[letterpaper, 10 pt, conference]{ieeeconf}  

\IEEEoverridecommandlockouts                              

\usepackage{tikz}
\def\*#1{\mathbf{#1}}

\usepackage{amsmath}
\usepackage{amsfonts}
\usepackage{dsfont}

\usepackage{booktabs} 
\usepackage{multirow}
\usepackage{graphicx}
\usepackage{cleveref}

\usepackage{amssymb}
\usepackage{pifont}

\usepackage{siunitx}

\title{\LARGE \bf
LS-VOS: Identifying Outliers in 3D Object Detections Using Latent Space Virtual Outlier Synthesis
}

\author{Aldi Piroli$^{1}$, Vinzenz Dallabetta$^{2}$, Johannes Kopp$^{1}$, Marc Walessa$^{2}$, \\ Daniel Meissner$^{2}$, and Klaus Dietmayer$^{1}$
\thanks{$^{1}$ Institute of Measurement, Control, and Microtechnology, Ulm University, Germany {\tt\small \{firstname.lastname\}@uni-ulm.de}}
\thanks{$^{2}$ BMW~AG, Petuelring 130, 80809~Munich,~Germany {\tt\small \{vinzenz.dallabetta, marc.walessa\}@bmw.de} and {\tt\small daniel.da.meissner@bmwgroup.com}}%
}

\newcommand\copyrighttext{%
	\footnotesize \copyright\,2023 IEEE. Personal use of this material is permitted. Permission from IEEE must be obtained for all other uses, in any current or future media, including reprinting/republishing this material for advertising or promotional purposes, creating new collective works, for resale or redistribution to servers or lists, or reuse of any copyrighted component of this work in other works.}%
\newcommand\copyrightnotice{%
	\begin{tikzpicture}[remember picture,overlay]%
	\node[anchor=south,yshift=10pt] at (current page.south) {\fbox{\parbox{\dimexpr\textwidth-2cm}{\copyrighttext}}};%
	\end{tikzpicture}%
	\vspace{-10pt}%
}

\begin{document}

\maketitle
\copyrightnotice
\thispagestyle{empty}
\pagestyle{empty}

\begin{abstract}
LiDAR-based 3D object detectors have achieved unprecedented speed and accuracy in autonomous driving applications.
However, similar to other neural networks, they are often biased toward high-confidence predictions or return detections where no real object is present.
These types of detections can lead to a less reliable environment perception, severely affecting the functionality and safety of autonomous vehicles.
We address this problem by proposing LS-VOS, a framework for identifying outliers in 3D object detections.
Our approach builds on the idea of Virtual Outlier Synthesis (VOS), which incorporates outlier knowledge during training, enabling the model to learn more compact decision boundaries.
In particular, we propose a new synthesis approach that relies on the latent space of an auto-encoder network to generate outlier features with a parametrizable degree of similarity to in-distribution features.
In extensive experiments, we show that our approach improves the outlier detection capabilities of a state-of-the-art object detector while maintaining high 3D object detection performance.
\end{abstract}

\section{Introduction}
\begin{figure}[t!]
    \centering
        \includegraphics[width=1\columnwidth]{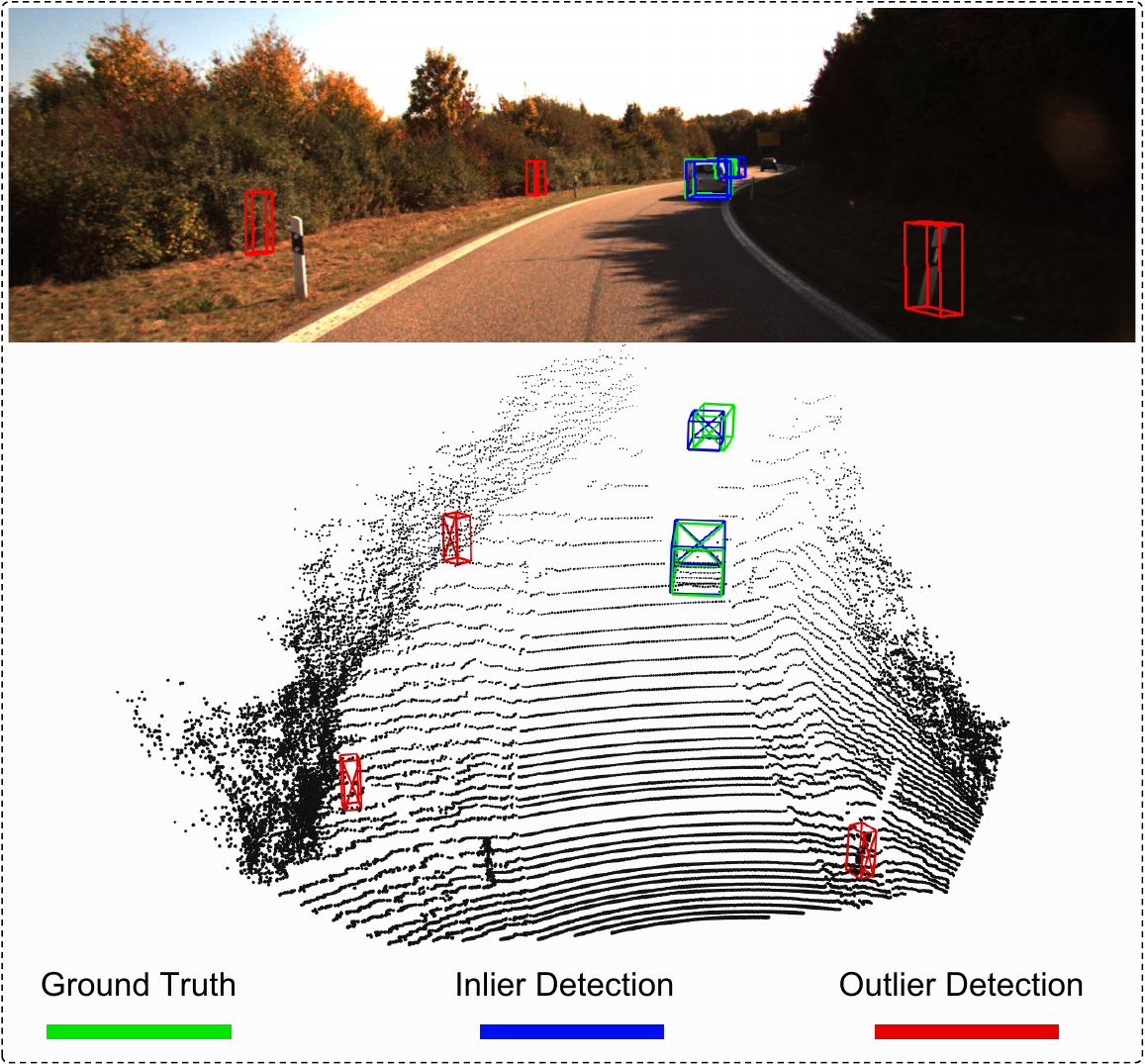}
    \caption{
Our proposed framework allows to identify outlier objects in 3D object detections by learning an uncertainty score during training for each predicted object. 
In the figure, we see that the detector correctly detects the two vehicles in the scene, but incorrectly detects roadside objects as pedestrians.
Using the learned confidence score, we can identify these false positive predictions.
The camera image is used for visualization only.
    }
    \label{Fig:teaser}
\end{figure}

LiDAR sensors are commonly used in autonomous driving applications to perceive the environment.
An important application is 3D object detection, which aims to localize and recognize the objects in a scene.
Although modern object detectors have achieved unprecedented performances, similar to other neural networks, they are often biased toward overconfident predictions~\cite{guo2017calibration, devries2018learning, Piroli2023Energy,hendrycks2018deep, liu2020energy,Piroli2022DetectionOC, hendrycks2016baseline, Piroli2022Robust3O}. 
One cause of this behavior is the use of only in-distribution (ID) target data during training, which can lead to the learning of open decision boundaries, resulting in overconfident predictions for unknown input data~\cite{du2022vos}. 
In the case of object detectors, this can lead to predicting out-of-distribution (OOD) objects as ID, or returning detections where no actual object of interest exists.
This last type of outlier prediction, often referred to as false positives (FP) or ghost detections, is the focus of this paper. 
These predictions can be extremely dangerous and problematic in safety-critical applications like autonomous driving. 
For example, predicting roadside vegetation as a pedestrian could cause the autonomous vehicle to make an emergency break, endangering its occupants and other road users.
An example of outlier object detections is shown in Fig.~\ref{Fig:teaser}. 

Although the topic of anomaly detection has gained interest in the image domain, little work has been done for 3D LiDAR point clouds. 
Feng et al.~\cite{feng2018towards} estimate the uncertainty of object detections by performing multiple forward passes.
However, the introduced latency in the system makes this approach unfeasible for real-time applications.
Hau et al.~\cite{hau2021shadow} distinguish between real and ghost objects by using a geometric approach to detect the presence of a shadow behind an object.
The shadow assumption works well for ghost objects caused by adversarial attacks, but it is often unfeasible to use in common real-world situations (e.g., clusters of objects or vehicles parked near buildings).
Recently, Huang et al.~\cite{huang2022out} have explored the use of features extracted from a detector backbone to distinguish between ID and OOD objects.
However, object detectors often incorrectly predict OOD objects due to the similarity of their features to ID objects, making these approaches less effective.

In this paper, we address the limitations mentioned above by proposing a framework for the identification of outliers in 3D object detections.
Our approach, dubbed Latent Space Virtual Outlier Synthesis (LS-VOS), builds upon the VOS framework~\cite{du2022vos}, which achieves state-of-the-art performances in outlier detection for image-based object detectors.
Specifically, we add an uncertainty head to the base structure of an object detector and train it to associate high scores with OOD objects and low scores with ID objects.
To generate outlier features, we first use an auto-encoder (AE) to learn the ID feature distribution.
Afterward, we synthesize outliers by adding noise during the AE reconstruction step of ID features.
Different from VOS, our approach does not make any prior assumption on the ID feature distribution and allows for the generation of outlier features with different degrees of similarity to ID features.
In extensive experiments on KITTI~\cite{geiger2013vision} and the Waymo Open Dataset~\cite{sun2020scalability}, we show that our approach improves the outlier discrimination ability of a state-of-the-art object detector~\cite{shi2023pv} while retaining high object detection performance.
A qualitative result of our framework is shown in Fig.~\ref{Fig:teaser}.

In summary, our main contributions are:
\begin{itemize}
    \item We propose LS-VOS, a framework for outlier identification in 3D object detections based on virtual outlier synthesis~\cite{du2022vos}.
    \item We present a novel approach to virtual outlier synthesis that makes no assumptions about the distribution of ID features and allows for a parametrizable degree of similarity between ID and generated OOD features.
    \item We test our method on real-world data and show that it improves outlier identification while retaining high 3D object detection performance.
\end{itemize}
\section{Related Work}
\subsection{3D Object Detection on Point Clouds}
3D object detectors aim to return a set of 3D bounding boxes containing the relevant objects in the scene and the corresponding object classes. 
To extract rich features from the input point cloud, different methods exist. 
PointNet++~\cite{qi2017pointnet++} processes the unordered point cloud directly without intermediate representation. 
VoxelNet~\cite{zhou2018voxelnet} first projects the points in a 3D voxel space and then uses full 3D convolutions to extract features.
SECOND~\cite{yan2018second} improves the voxel-based architecture by using sparse convolutions, greatly reducing computation times.
Modern object detectors like PV-RCNN~\cite{shi2020pv} and PV-RCNN++~\cite{shi2023pv} use a combination of point-based and voxel-based feature extraction to achieve state-of-the-art results.

\subsection{Anomaly Detection Methods}
A large body of work can be found for deep learning-based anomaly detections. 
Hendrycks et al.~\cite{hendrycks2016baseline} propose to use the output softmax score of a network to detect if an input is misclassified or OOD. 
They find that the softmax score is biased towards high confidence values even for random input noise.
DeVries et al.~\cite{devries2018learning} propose learning the confidence of a prediction in the context of image classification using an auxiliary branch. 
The confidence output can be used to detect OOD samples by choosing an appropriate threshold. 
Hendrycks et al.~\cite{hendrycks2018deep} propose the outlier exposure method, which uses an auxiliary dataset of outliers to train the network to discriminate between inliers and outliers. 
Liu et al.~\cite{liu2020energy} use the energy function to map the output classification logits of a network to a single real number, called energy score. 
Compared to the standard softmax~\cite{hendrycks2016baseline}, the energy score is shown to be less prone to overconfident predictions.
Zhang et al.~\cite{zhang2023mixture} propose an outlier synthesis method for fine-grained classes based on the MixUp~\cite{zhang2017mixup} and CutMix~\cite{yun2019cutmix} operations between ID and OOD features.
Du et al.~\cite{du2022vos} propose the virtual outlier synthesis (VOS) method, which allows the generation of outlier samples in the feature space.
They model the features of ID objects extracted by a 2D object detector backbone as class-conditional normal distributions and sample outlier features from low-likelihood regions.
These outliers are then used to train an additional uncertainty head which enables the detection of outlier objects.
Our proposed LS-VOS builds upon the VOS framework and applies the idea of outlier knowledge incorporation during the training of an object detector.
In addition, LS-VOS reduces the restrictive assumption of the normally distributed ID features by learning the feature distributions using an AE network.

Few works exist for anomaly identification in 3D object detection~\cite{bogdoll2022anomaly}.
Feng et al.~\cite{feng2018towards} use Monte Carlo Dropout and Deep Ensembles to estimate the epistemic uncertainty in the category classification.
These approaches rely on multiple forward passes (up to $40$) of the object detector, making it infeasible for real-time applications like autonomous driving.
Hau et al.~\cite{hau2021shadow} propose the Shadow-catcher framework to detect ghost objects in LiDAR-based 3D object detectors caused by an adversarial physical attack.  
Based on this observation, it is possible to derive an anomaly score for each object, which can be used to distinguish between real and ghost predictions.
Huang et al.~\cite{huang2022out} explore the use of classification OOD detection methods for 3D object detections.
They construct point clouds with OOD objects by inserting objects extracted from a database of synthetic and real point clouds which do not appear in the target dataset. 
To differentiate between ID and OOD detections, they use the Mahalanobis distance~\cite{lee2018simple} between the features associated with a detected object and the ID feature distribution.
Similarly, they use the log-likelihood of an object's feature as an uncertainty score, applying normalizing flows for the ID feature distribution estimation~\cite{dinh2016density}.
\section{Method}
\begin{figure*}[t!]
    \centering
        \includegraphics[width=1\textwidth]{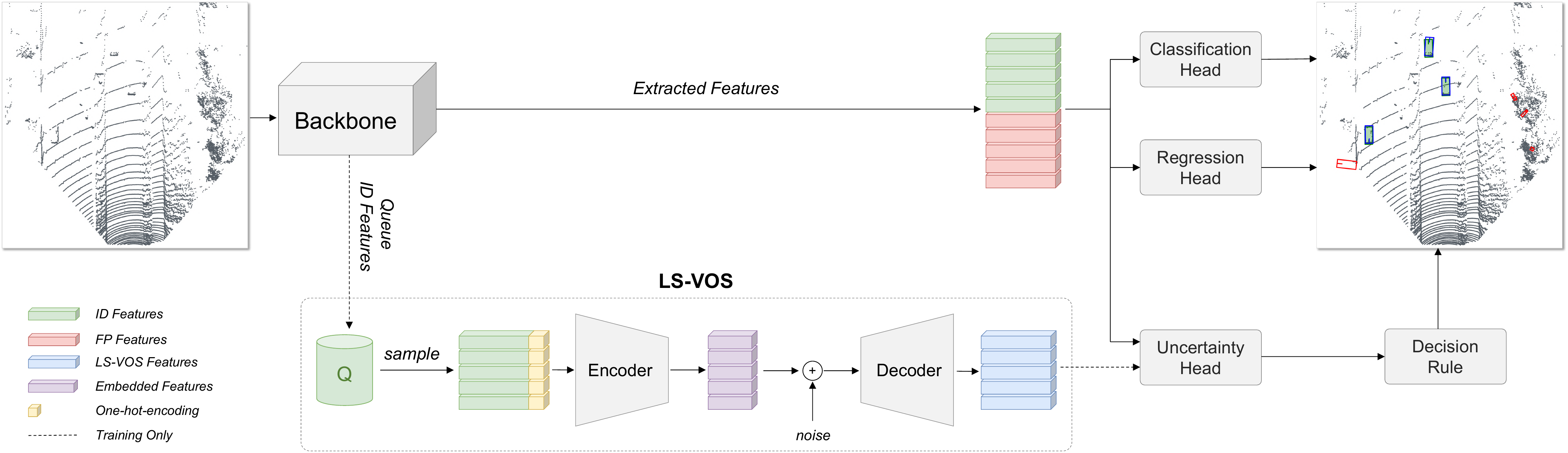}
    \caption{
Overview of the proposed LS-VOS framework.
During training, the backbone of a 3D object detector extracts features from the input point cloud. 
Based on the IoU with the ground truth bounding boxes, features are divided into in-distribution (ID) and false positives (FP).
The ID features, concatenated with the one-hot-encoded predicted class, are saved at each training step in a queue Q.
To synthesize virtual outliers, ID features are randomly sampled  from Q and an encoder (E) is used to embed them.
Random noise is added to the embedded features which are then reconstructed using a decoder (D).
These reconstructed features (virtual outliers) are used together with ID and FP to train the uncertainty head, which returns high scores for OOD objects and low scores for ID objects. 
In the detection plot, we use the color green for ground truth boxes and blue for predictions.
We use red to highlight the predicted boxes classified as OOD.}
    \label{Fig:method}
\end{figure*}
\label{sec:method}
In the following, we describe our proposed LS-VOS framework, which aims to discriminate between inlier and outlier 3D object detections. 
An overview of the method is given in Fig.~\ref{Fig:method}.

\subsection{Latent Space Virtual Outlier Synthesis (LS-VOS)}
\textbf{Feature Extraction.} 
In general, given a point cloud $\*p \in \mathbb{R}^{N \times C}$ composed of $N$ total points with $C$ features each, an object detector $f$ extracts $\*u\in \mathbb{R}^{M \times D}$ features using a backbone component $f_\text{back}$, where $M$ is the number of detection proposals and $D$ the dimension of each feature vector.
These features are then processed by a classification head $f_\text{cls}$, which classifies each feature into one of the $K$ ID classes, and a regression head $f_\text{reg}$ which returns a bounding box containing the detected object. 
The predicted objects can be divided into true positive (TP) and false positive (FP) predictions, depending on the intersection-over-union (IoU) between the ground truth bounding boxes and the regressed ones.
Based on this classification, we can split the extracted features into $\*u_\text{ID} \in  \mathbb{R}^{M_\text{ID} \times D}$ (from TP predictions) and  $\*u_\text{FP}  \in  \mathbb{R}^{M_\text{FP} \times D}$ (from FP predictions) with $M = M_\text{ID} + M_\text{FP}$.

\textbf{Feature Density Estimation.} 
We now aim to estimate the density distribution of the ID features.
For this purpose, we use an auto-encoder $\phi = d\big(e(\*x)\big):\mathbb{R}^{M'_\text{ID} \times (D+K)} \rightarrow \mathbb{R}^{M'_\text{ID} \times (D)}$, where $e$ is the encoder and $d$ the decoder. 
For each feature in $\*u_\text{ID}$ extracted during training, we augment it with the one-hot-encoded predicted class and store it in a FIFO queue $\mathrm{Q}$ of fixed size. 
We then randomly sample $M'_\text{ID}$ features $\*x \in \mathbb{R}^{M'_\text{ID} \times (D+K)} $ from $\mathrm{Q}$ and use them as training data for $\phi$.
In particular, we first encode the input features $\*z = e(\*x)$ with $\*z \in \mathbb{R}^{M'_\text{ID} \times D'}$ and $D' < D+K$.
Then, the features are reconstructed using  $\tilde{\*x} = d(\*z)$ with $\tilde{\*x} \in \mathbb{R}^{M'_\text{ID} \times D}$.
We train $\phi$ in an end-to-end manner together with the detector $f$ using the mean squared error loss function
\begin{align}\label{eq:auto_encoder_loss}
  \mathcal{L}_\text{AE} = 
   \mathbb{E}_{\*x \sim \mathrm{Q}} &\left[ \big(\*x- \phi(\*x)\big)^2\right].
\end{align}


\textbf{LS-VOS.}
Our method for the synthesis of virtual outliers consists of reconstructing $\*u_\text{ID}$ features using $\phi$, with the addition of noise in the reconstruction step.
This simple approach allows the synthesis of outlier features with varying degrees of similarity to the input features. 
In particular, given the vector of ID features $\*u_\text{ID}$ we get $\*u'_\text{ID} \in \mathbb{R}^{M_\text{ID} \times (D+K)}$ by concatenating the one-hot-encoded predicted classes and encode it using $\*z_\text{ID} = e(\*u_\text{ID})$.
We then derive a noise vector $\*o = \beta \cdot(\alpha + \*o' )  \in\mathbb{R}^{M_\text{ID} \times D'}$ where $\*o'$ is sampled from an uniform distribution $\mathcal{U}(0, 1) $ and $\alpha$, $\beta \in \mathbb{R}$.
Afterward, we add the noise vector to the encoded features resulting in $\*z_\text{ID}^* = \*z_\text{ID} + \*o$.
Finally, we reconstruct the noisy features and obtain the virtual outliers  $\*v = d(\*z_\text{ID}^*) \in \mathbb{R}^{M_\text{ID}\times D}$.
The values of $\alpha$ and $\beta$ can be used as a parameter to determine the distance between the input ID features and the reconstructed ones. 
In contrast to VOS, which samples virtual outliers only from low-likelihood regions of the ID distribution, we generate outliers that are near the ID input features (small noise values) and far away from them (larger noise values).
The model can then learn from both \textit{easy} and \textit{hard} to distinguish examples.
In Fig.~\ref{Fig:tSNE_plots} we show t-SNE plots of ID and LS-VOS features.

\subsection{Outlier Detection}
As mentioned in the previous section, object detectors are usually composed of a classification and a regression head.
The confidence of each detection is derived by applying a scoring function to the output logits of the $f_\text{cls}$ head.
The maximum softmax probability is usually used, which allows for a probabilistic interpretation of the box confidence.
To determine if a predicted object is ID or OOD, we augment the base structure of an object detector by adding an additional uncertainty estimation head $f_\text{unc}(\*u)$.

During training, the uncertainty estimation head  $f_\text{unc}(\*u):\mathbb{R}^{M_\text{train}\times D} \rightarrow \mathbb{R}^{M_\text{train}}$ is trained using the binary sigmoid loss function
\begin{align}\label{eq:ood_head_loss}
\begin{split}
   \mathcal{L}_\text{uncertainty} = 
   \mathbb{E}_{\*u \sim \*u_\text{OOD}} &\left[  -\frac{1}{1+\exp^{-f_\text{unc}(\*u)}} \right] 
  + \\  \mathbb{E}_{\*u \sim \*u_\text{ID}}& \left[  -\frac{\exp^{-f_\text{unc}(\*u)}}{1+\exp^{-f_\text{unc}(\*u)}} \right],
  \end{split}
\end{align}
where $\*u_\text{OOD} = [\*v, \*u_\text{FP}] \in \mathbb{R}^{M_\text{OOD} \times D}$ is the vector containing both synthesized virtual outliers and FP features, with $M_\text{train} = M_\text{ID} + M_\text{OOD}$.
By minimizing the loss~\eqref{eq:ood_head_loss}, the uncertainty estimation head learns to associate low scores with ID features and high scores with OOD ones. 
The total loss function is then:
\begin{align}
\label{eq:total_loss}
  \mathcal{L}_\text{tot} = \mathcal{L}_\text{det} + \mathcal{L}_\text{AE} + \lambda \mathcal{L}_\text{uncertainty},  
\end{align}
where $\mathcal{L}_\text{det}$ is the default loss function of the object detector and $\lambda$ a weighting parameter.

During inference, the outlier detection head $f_\text{unc}(\*u):\mathbb{R}^{M\times D} \rightarrow \mathbb{R}^{M}$ returns a score for each of the predicted objects.
In applications where a hard classification between ID and OOD detections is required, a decision rule can be used:
\begin{align}
\label{eq:decision_rule}
g_i(\*u; \tau; f_\text{unc}) =
\begin{cases}
\text{ID} & \quad \text{if } f_{\text{unc}, i}(\*u) \leq \tau, \\
\text{OOD}  & \quad \text{else},
\end{cases}
\end{align}
where $f_{\text{unc}, i}(\*u)$ is the output corresponding to the $i$-th feature and $i=1, \dots, M$. 
Here $\tau$ is a threshold parameter that can be chosen appropriately depending on the application at hand.
For example, in autonomous driving applications, one can choose a threshold for which a high  number (e.g., $95\%$) of ID predictions are correctly classified.
\section{Experiments}

\begin{table*}[t!]
    \centering
    \caption{Outlier detection results on the KITTI and the Waymo Open Dataset. All values are in percentage. $\uparrow$ means higher values are better and $\downarrow$ means lower values are better. Bold numbers are the superior results.}
    \resizebox{0.8\textwidth}{!}{%
        \begin{tabular}{@{}lccclccc@{}}
            \toprule
            \multirow{2}{*}{\textbf{Method}}         & \multicolumn{3}{c}{\textbf{KITTI }} &                          & \multicolumn{3}{c}{\textbf{Waymo Open Dataset}}                                                                                         \\ \cmidrule(lr){2-4} \cmidrule(l){6-8}
                                                     & \textbf{AUROC}~$\uparrow$           & \textbf{AUPR}~$\uparrow$ & \textbf{FPR95}~$\downarrow$        &  & \textbf{AUROC}~$\uparrow$ & \textbf{AUPR}~$\uparrow$ & \textbf{FPR95}~$\downarrow$ \\ \midrule
            Default score                            & 94.04                               & 96.27                    & 32.34                              &  & 92.52                     & 91.76                    & 40.08                       \\
            Shadow-catcher~\cite{hau2021shadow}      & 78.94                               & 83.35                    & 73.31                              &  & 71.18                     & 63.48                    & 89.63                       \\
            Mahalanobis distance~\cite{huang2022out} & 85.76                               & 86.49                    & 40.99                              &  & 86.07                     & 78.80                    & 49.68                       \\
            Normalizing flows~\cite{huang2022out}    & 92.22                               & 95.46                    & 41.44                              &  & 92.33                     & 91.15                    & 38.43                       \\
            LS-VOS (ours)                            & \textbf{95.50}                      & \textbf{97.91}           & \textbf{26.47}                     &  & \textbf{94.69}            & \textbf{95.87}           & \textbf{32.83}              \\ \bottomrule
        \end{tabular}
    }
    \label{tab:ood_detection}
\end{table*}

\subsection{Experiments Setup}
\textbf{Datasets and Evaluation Metrics.}
We evaluate our proposed method on the KITTI dataset~\cite{geiger2013vision}, which contains $3712$  training samples and $3769$ validation samples.
We also evaluate on the Waymo Open Dataset (WOD)~\cite{sun2020scalability}, which is currently one of the largest dataset for autonomous driving applications containing approximately $\SI{158}{\kilo{}}$ training and $\SI{40}{\kilo{}}$ validation samples. 

Following prior work on OOD detection~\cite{hendrycks2016baseline, devries2018learning, hendrycks2018deep, liu2020energy, du2022vos, zhang2023mixture, huang2022out}, we use as evaluation metrics the receiver operating characteristic curve (AUROC), the area under the precision-recall curve (AUPR), the FPR95, which measures the false positive rate when the true positive rate is equal to $95\%$.
Additionally, for the methods which require the fine-tuning of the object detector, we report the expected calibration error (ECE), which measures how well the predicted probabilities match the ground truth probability distribution.
To evaluate 3D object detection performances, we use the dataset's official metrics, which are the bird's-eye view (BEV) and 3D average precision (AP) for the KITTI dataset and the average precision weighted by heading (APH) for the WOD.
Given the set of predicted bounding boxes, we label as OOD all the boxes which have IoU with the ground truth bounding boxes smaller than a threshold, which is $0.7$ for \textit{vehicles} and $0.5$ for \textit{pedestrians} and \textit{cyclists}.
These IoU thresholds are the same ones used for both KITTI and WOD to compute the AP metrics.

\begin{figure}[t!]
    \centering
        \includegraphics[width=0.98\columnwidth]{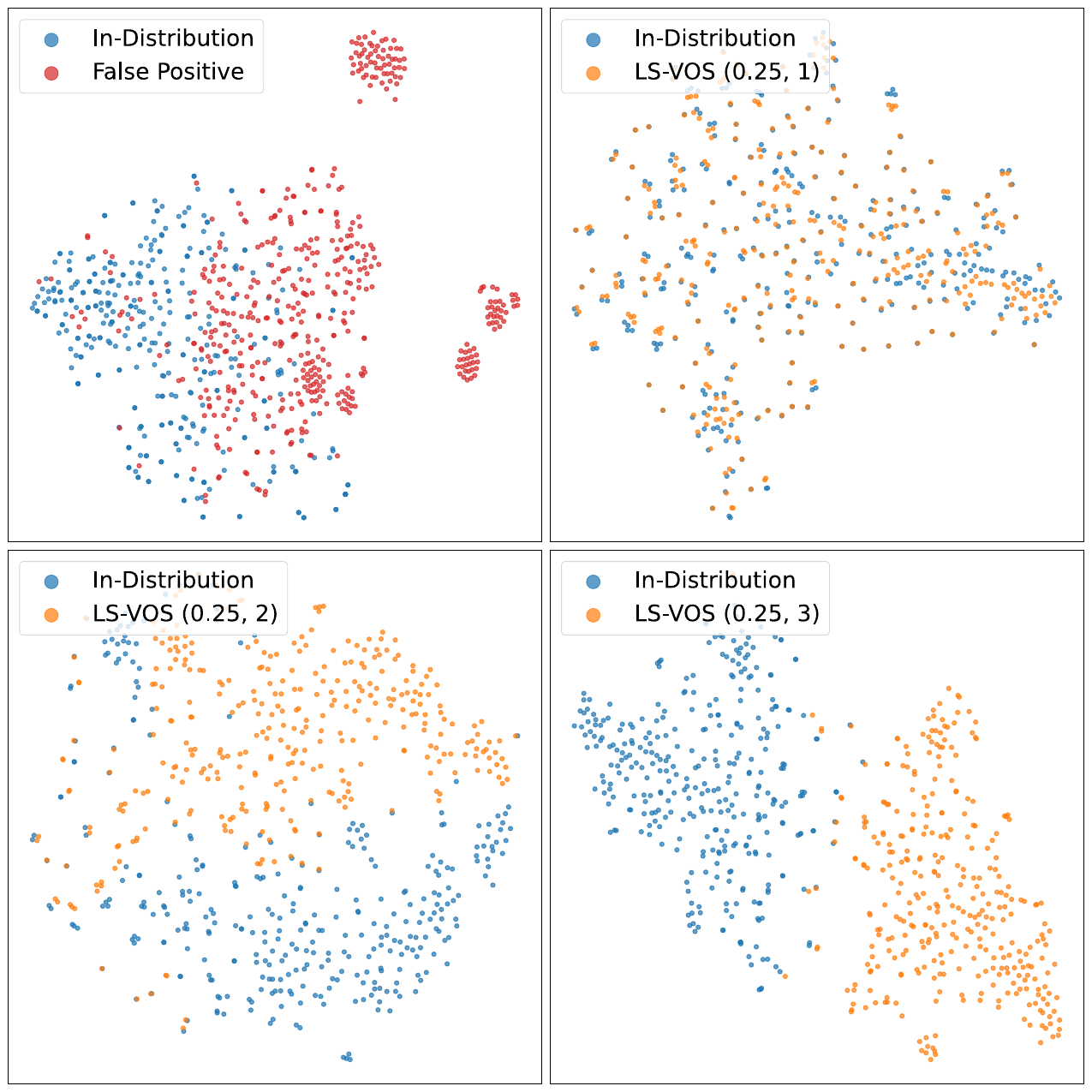}
    \caption{
t-SNE visualization of In-Distribution (ID) and false positives features (top-left), ID and LS-VOS features with parameters ($\alpha$, $\beta$) (top-right, bottom-left, bottom-right).  
    }
    \label{Fig:tSNE_plots}
\end{figure}

\textbf{Baselines Details.}
We compare our proposed method against other approaches for ghost object detection~\cite{hau2021shadow} (Shadow-catcher), and OOD detection in 3D object detectors~\cite{huang2022out} (Mahalanobis distance and normalizing flows).
In particular, we re-implement Shadow-catcher using the 3D shadow region proposal with uniform shadow height of $0.2$$\si{\metre}$ and use the shadow region anomaly score as outlier score.
For the normalizing flows method, we use RealNVP~\cite{dinh2016density} with $32$ normalizing flows layers and a hidden dimension of $1024$.
We also compare to other state-of-the-art virtual outlier synthesis methods like VOS~\cite{du2022vos} and Outlier Mixture Exposure~\cite{zhang2023mixture}.
For the latter, we apply the proposed LinearMix operation with weight parameter $0.5$ between ID and FP features to generate OOD samples.
Similar to~\cite{du2022vos}, we also report the results of outlier synthesis performed by randomly sampled noise from $\mathcal{N}(0,1)$ (random noise) and random noise sampled from $\mathcal{U}(0, 1)$ and added to ID features (noisy ID features).
As object detector for all of our experiments, we use PV-RCNN++~\cite{shi2023pv}, a high-performing state-of-the-art detector, and use the official implementation provided in~\cite{openpcdet2020}.
The confidence score of PV-RCNN++ is used as baseline (default score). 
For all the methods which operate in the feature space (Mahalanobis distance, normalizing flows, LinearMix, LSVOS, $\dots$), we use the features extracted after the RoI-grid pooling module of PV-RCNN++.

\textbf{Method Implementation Details.}
For the encoder $e$, we use fully connected layers of dimensions $256, 128, 128$, the first two with ReLU activation functions.
Similarly, for the decoder $d$, we use fully connected layers of dimensions $128, 256, 256$, the first two with ReLU activation functions.
Therefore, the used AE compresses the input features from a dimension of $255$ to $128$.
For the $f_\text{unc}$ head, we use fully connected layers of dimension $256, 256, 1$, and use ReLU activations for the first two.
The output of $f_\text{unc}$ is used as anomaly score in all experiments.  

For each class $k=1, \dots, K$, we save in the feature queue $\mathrm{Q}$ a maximum of $1000$ features and during training use sampling size of $500$ for each class, resulting in $M' = 500\cdot K$ total training features.
We first train the AE for $50$ and $5$ epochs for KITTI and WOD respectively, using the loss function~\eqref{eq:total_loss} with $\lambda=0$, Adam optimizer~\cite{kingma2014adam}, and a constant learning rate of $1\times10^{-3}$.
Afterward, we continue training with $\lambda=1$ for $20$ and $10$ epochs on the KITTI and WOD datasets using the same optimization parameters.
Unless otherwise stated, we use as noise sampling parameters $\alpha=0.25$ and $\beta=1$ for all experiments.

\subsection{Outlier Detection Results}
\begin{figure*}[t!]
    \centering
        \includegraphics[width=0.92\textwidth]{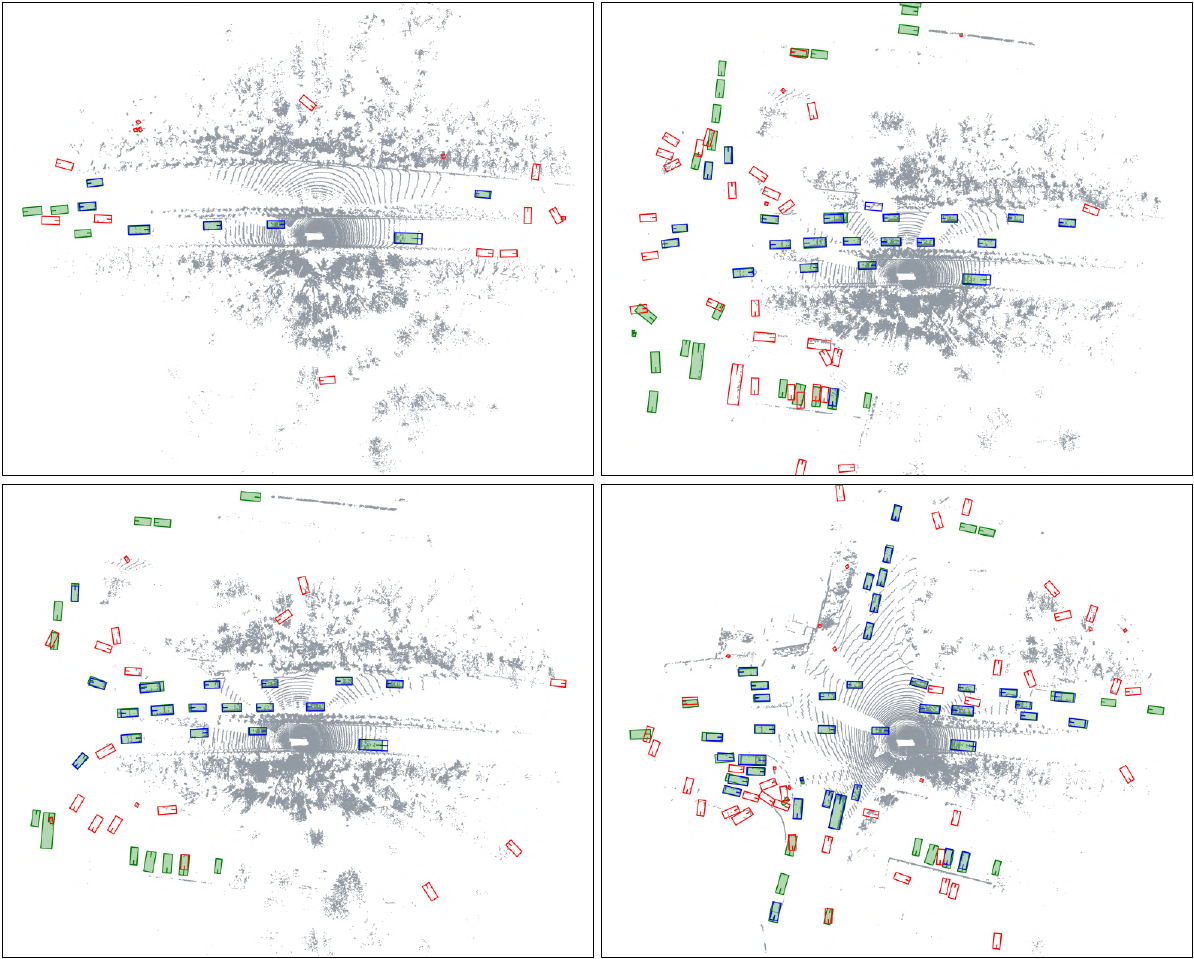}
    \caption{
Qualitative results of PV-RCNN++ trained with the proposed LS-VOS framework.
All four scenes are from the Waymo Open Dataset validation set.
In green, we report the ground truth bounding boxes, and in blue the predicted bounding boxes.
In red, we highlight the detections classified as OOD using the decision rule described in~\eqref{eq:decision_rule} with threshold $\tau$ chosen at $95\%$ true positive rate.
}    \label{Fig:qualitative_results}
\end{figure*}

We start the evaluation by comparing the OOD detection performances of LS-VOS against other state-of-the-art approaches.
As shown in Table~\ref{tab:ood_detection}, LS-VOS outperforms both the baseline score and the other approaches by a large margin. 
Compared to PV-RCNN++ default score, training with LS-VOS improves the FPR95 score by $5.87\%$ points on KITTI and $7.25\%$ points on the WOD.
This shows that the score learned by $f_{\text{unc}}$ performs better in distinguishing between ID and OOD predictions than the default score. 
LS-VOS also has higher performances than the other state-of-the-art approaches on both KITTI and WOD. 
For example, compared to normalizing flows~\cite{huang2022out} the performance on WOD improves by $2.36\%$ AUROC, $4.72\%$ AUPR, and $5.6\%$ FPR95 points.
Methods like the Mahalanobis distance and normalizing flows rely on the distance of an object feature to the ID feature distribution to determine whether it is ID or OOD.
However, as we can see from Fig.~\ref{Fig:tSNE_plots}, a significant number of FP features are closely clustered to the ID features. 
Consequently, this proximity diminishes the efficacy of methods in distinguishing between the two types of samples.
In Fig.~\ref{Fig:teaser} and~\ref{Fig:qualitative_results} we report some qualitative examples of PV-RCNN++ trained with LS-VOS.

In Table~\ref{tab:det3d_kitti} and~\ref{tab:det3d_waymo} we report the effect on 3D object detection performances of PV-RCNN++ vanilla and trained with LS-VOS.
We observe that when training PV-RCNN++ with LS-VOS on KITTI, the performance on the Pedestrian class improves by $3.6\%$ BEV AP and $4.05\%$ 3D AP.
However, we also see a small degradation in performance for the Car and Cyclist classes.
In the WOD, which is much larger in size, the performance of the baseline and LS-VOS trained detector are similar.
\begin{table}[t!]
    \centering
    \caption{Effect of LS-VOS on the 3D object detection performances in the KITTI dataset.
    The results are evaluated on the moderate difficulty level and reported as BEV/3D AP.    
    All values are in percentage.}
    \resizebox{1\columnwidth}{!}{%
\begin{tabular}{@{}lcccc@{}}
\toprule
\textbf{Method} & \textbf{Car} & \textbf{Pedestrian} & \textbf{Cyclist} & \textbf{Avg. (BEV/3D AP)}~$\uparrow$ \\ \midrule
Baseline       & 90.74~/~84.58  & 59.40~/~55.75          & 74.71~/~73.16      & 74.95~/~71.16   \\
LS-VOS          & 90.10~/~83.23   & 63.00~/~59.80             & 74.24~/~71.83      & 75.78~/~71.62   \\ \bottomrule
\end{tabular}
}
\label{tab:det3d_kitti}
\end{table} 

\begin{table}[t!]
    \centering
    \caption{Effect of LS-VOS on the 3D object detection performances on the Waymo Open Dataset.
    The results are evaluated on two levels of difficulty, with LEVEL 2 being cumulative and including LEVEL 1.    
    All values are in percentage.}
    \resizebox{0.95\columnwidth}{!}{%
\begin{tabular}{@{}llcccc@{}}
\toprule
\textbf{Difficulty}      & \textbf{Method} & \textbf{Vehicle} & \textbf{Pedestrian} & \textbf{Cyclist} & \textbf{Average} \\ \midrule
\multirow{2}{*}{LEVEL 1} & Baseline   & 75.92            & 65.81               & 70.37            & 70.70         \\
                         & LS-VOS          & 75.52            & 65.87               & 70.38            & 70.59         \\ \midrule
\multirow{2}{*}{LEVEL 2} & Baseline   & 67.40            & 57.98               & 67.78            & 64.39         \\
                         & LS-VOS          & 67.02            & 58.11               & 67.79            & 64.31         \\ \bottomrule
\end{tabular}
}
\label{tab:det3d_waymo}
\end{table}

\subsection{Comparison With Other Outlier Synthesis Methods}
In Table~\ref{tab:vos_methods} we compare LS-VOS with other state-of-the-art methods for outlier synthesis.
LS-VOS outperforms all of the other methods in terms of AUROC, AUPR, and FPR95. 
Compared to VOS, we improve the AUROC, AUPR, and FPR95 scores by $0.43\%$, $0.48\%$, and $1.88\%$ points respectively, while maintaining similar 3D object detection performances.
We note that an apparent side effect of virtual outlier synthesis training is an increase in ECE, with almost all of the tested methods resulting in higher values than the default prediction score. 

\begin{table}[t!]
    \centering
    \caption{Comparison of virtual outlier synthesis methods trained and evaluated on the KITTI dataset.
        AP results are reported for the moderate difficulty level.
        $\uparrow$ means higher values are better and $\downarrow$ means lower values are better. Bold numbers are the superior results.}
    \resizebox{1\columnwidth}{!}{%

        \begin{tabular}{@{}lccccc@{}}
            \toprule
            \textbf{Method}                   & \textbf{AUROC}~$\uparrow$ & \textbf{AUPR}~$\uparrow$ & \textbf{FPR95}~$\downarrow$ & \textbf{Avg. (BEV/3D AP)}~$\uparrow$ & \textbf{ECE}~$\downarrow$ \\ \midrule
            Default score                     & 94.04                     & 96.27                    & 32.34                       & 74.95~/~71.16                        & 0.28                      \\
            Random noise                      & 94.78                     & 97.14                    & 29.00                       & 75.08~/~71.31                        & 0.31                      \\
            Noisy ID features                 & 93.79                     & 95.92                    & 33.23                       & 74.94~/~71.44                        & \textbf{0.26}             \\
            LinearMix~\cite{zhang2023mixture} & 95.08                     & 97.53                    & 27.91                       & 74.84~/~70.56                        & 0.32                      \\
            VOS~\cite{du2022vos}              & 95.07                     & 97.43                    & 28.35                       & 75.52~/~\textbf{71.65}               & 0.31                      \\
            LS-VOS (ours)                     & \textbf{95.50}            & \textbf{97.91}           & \textbf{26.47}              & \textbf{75.78}~/~71.62               & 0.35                      \\ \bottomrule
        \end{tabular}
    }
    \label{tab:vos_methods}
\end{table}

\subsection{Ablation Studies}
\textbf{Noise Parameters.}
In Table~\ref{tab:noise_ablation} we report the impact of different noise sampling parameters on performance.
The parameter $\alpha$ determines the smallest possible noise value, whereas $\beta$ affects the magnitude of the added noise. 
We observe that having a high magnitude in the noise sampling ($\alpha=0.25, \beta=10$) is beneficial to OOD detection, suggesting that training with a mix of easy and challenging examples can enhance the model's ability to detect OOD objects.
Both low and high magnitude values lead to high ECE, making moderate values better suited to maintain a lower calibration error.  
The parameter $\alpha$ is also important for both OOD and 3D object detection performances.
For example, when comparing $\alpha=0$ and $\alpha=0.25$ at $\beta=1$, we see that although $\alpha=0$ has better OOD detection, both the BEV and 3D AP performances are lower. 
When using small $\alpha$ values it can happen that the resulting virtual outliers are almost indistinguishable from ID features, making it challenging for the uncertainty head to learn to distinguish between them.

\textbf{Outlier Loss Weight.}
In Table~\ref{tab:weight_loss_ablation}, we evaluate the impact of the loss weight parameter $\lambda$ on the model's performance. 
Higher values of $\lambda$ improve OOD detection performance but also negatively affect the ECE.
Conversely, lower values, such as $\lambda = 0.1$, yield better OOD detection than the baseline, while maintaining comparable object detection performance and ECE.
\begin{table}[t!]
    \centering
    \caption{Effect of different noise parameters on LS-VOS trained and evaluated on the KITTI dataset. AP results are reported for the moderate difficulty level.
    $\uparrow$ means higher values are better and $\downarrow$ means lower values are better. Bold numbers are the superior results.}
    \resizebox{1\columnwidth}{!}{%
\begin{tabular}{@{}ccccccc@{}}
\toprule
\textbf{$\alpha$} & \textbf{$\beta$} & \textbf{AUROC}~$\uparrow$ & \textbf{AUPR}~$\uparrow$  & \textbf{FPR95}~$\downarrow$ & \textbf{Avg. (BEV/3D AP)}~$\uparrow$ & \textbf{ECE}~$\downarrow$  \\ \midrule
0.0        & 0.1           & 96.05          & 98.49          & 24.13          & 75.38~/~70.30            & 0.40          \\
0.0        & 0.5           & \textbf{96.23} & \textbf{98.61} & 23.64          & 75.66~/~70.72            & 0.40          \\
0.0        & 1.0           & 95.61          & 98.23          & 24.00          & 74.82~/~69.76            & \textbf{0.34} \\
0.25        & 1.0           & 95.50          & 97.91          & 26.47          & \textbf{75.78}~/~\textbf{71.62}           & 0.35          \\
0.25        & 5.0           & 95.47          & 97.97          & 25.52          & 74.37~/~70.40            & 0.36          \\
0.25        & 10.0          & 95.93          & 98.38          & \textbf{23.37} & 74.57~/~70.69            & 0.38          \\ \bottomrule
\end{tabular}
}
\label{tab:noise_ablation}
\end{table} 
\begin{table}[t!]
    \centering
    \caption{Effect of the outlier loss weight $\lambda$ parameters on LS-VOS trained and evaluated on the KITTI dataset.
        AP results are reported for the moderate difficulty level.
        $\uparrow$ means higher values are better and $\downarrow$ means lower values are better. Bold numbers are the superior results.}
    \resizebox{0.95\columnwidth}{!}{%
        \begin{tabular}{@{}cccccc@{}}
            \toprule
            \textbf{$\lambda$} & \textbf{AUROC}~$\uparrow$ & \textbf{AUPR}~$\uparrow$ & \textbf{FPR95}~$\downarrow$ & \textbf{Avg. (BEV/3D AP)}~$\uparrow$ & \textbf{ECE}~$\downarrow$ \\ \midrule
            0.1                & 94.27                     & 96.61                    & 30.73                       & 75.05~/~70.95                        & \textbf{0.29}             \\
            0.5                & 94.96                     & 97.48                    & 27.66                       & 75.21~/~70.66                        & 0.33                      \\
            1.0                & 95.50                     & 97.91                    & 26.47                       & 75.78~/~71.62                        & 0.35                      \\
            2.0                & 96.30                     & 98.58                    & 22.14                       & \textbf{76.13}~/~\textbf{72.11}      & 0.39                      \\
            5.0                & \textbf{96.59}            & \textbf{98.78}           & \textbf{20.58}              & 75.34~/~71.23                        & 0.41                      \\ \bottomrule
        \end{tabular}
    }
    \label{tab:weight_loss_ablation}
\end{table}

\subsection{Discussion}
Virtual outlier synthesis approaches such as LS-VOS, VOS, and Feature Mixup improve the performance of OOD detection while maintaining high 3D object detection performances.
In our experiments, however, we see that this comes at the cost of an increase in the ECE.
One possible explanation for this effect is that more compact decision boundaries learned by the model lead to unintended lower confidence values for ID features.
A possible solution for the mitigation of this effect can be the use of post-training calibration methods like temperature scaling~\cite{guo2017calibration}, or the use of a lower weighting value $\lambda$ as showed in Table~\ref{tab:weight_loss_ablation}.
Additionally, we notice that models trained with outlier synthesis tend to predict a higher number of objects per scan. 
This effect might require a more restrictive value of the parameter $\tau$ described in~\eqref{eq:decision_rule}, resulting in extreme cases in the classification of ID objects as OOD.  

\section{Conclusion}
In this paper, we present LS-VOS, a framework for the identification of outliers in 3D object detections.
Our approach is based on the VOS framework, which incorporates outlier knowledge during training allowing a detector to learn more compact decision boundaries.
Building on this idea, we propose a novel outlier synthesis method based on the modeling of in-distribution feature distribution using an auto-encoder (AE) network. 
Our synthesis approach consists of adding noise during the (AE) reconstruction step, allowing to generate both easy and difficult examples in a parametrizable way.
Experiments on the KITTI and Waymo Open Dataset show that our framework improves the outlier object detection capabilities of an object detector, while maintaining similar 3D object detection performance.

\bibliographystyle{IEEEtran}
\bibliography{mybib}

\begin{thebibliography}{10}
\providecommand{\url}[1]{#1}
\csname url@rmstyle\endcsname
\providecommand{\newblock}{\relax}
\providecommand{\bibinfo}[2]{#2}
\providecommand\BIBentrySTDinterwordspacing{\spaceskip=0pt\relax}
\providecommand\BIBentryALTinterwordstretchfactor{4}
\providecommand\BIBentryALTinterwordspacing{\spaceskip=\fontdimen2\font plus
\BIBentryALTinterwordstretchfactor\fontdimen3\font minus
  \fontdimen4\font\relax}
\providecommand\BIBforeignlanguage[2]{{%
\expandafter\ifx\csname l@#1\endcsname\relax
\typeout{** WARNING: IEEEtran.bst: No hyphenation pattern has been}%
\typeout{** loaded for the language `#1'. Using the pattern for}%
\typeout{** the default language instead.}%
\else
\language=\csname l@#1\endcsname
\fi
#2}}

\bibitem{guo2017calibration}
C.~Guo, G.~Pleiss, Y.~Sun, and K.~Q. Weinberger, ``On calibration of modern
  neural networks,'' in \emph{International conference on machine
  learning}.\hskip 1em plus 0.5em minus 0.4em\relax PMLR, 2017, pp. 1321--1330.

\bibitem{devries2018learning}
T.~DeVries and G.~W. Taylor, ``Learning confidence for out-of-distribution
  detection in neural networks,'' \emph{arXiv preprint arXiv:1802.04865}, 2018.

\bibitem{Piroli2023Energy}
A.~Piroli, V.~Dallabetta, J.~Kopp, M.~Walessa, D.~Meissner, and K.~Dietmayer,
  ``Energy-based detection of adverse weather effects in lidar data,''
  \emph{IEEE Robotics and Automation Letters}, vol.~8, no.~7, pp. 4322--4329,
  2023.

\bibitem{hendrycks2018deep}
D.~Hendrycks, M.~Mazeika, and T.~Dietterich, ``Deep anomaly detection with
  outlier exposure,'' \emph{arXiv preprint arXiv:1812.04606}, 2018.

\bibitem{liu2020energy}
W.~Liu, X.~Wang, J.~Owens, and Y.~Li, ``Energy-based out-of-distribution
  detection,'' \emph{Advances in Neural Information Processing Systems},
  vol.~33, pp. 21\,464--21\,475, 2020.

\bibitem{Piroli2022DetectionOC}
A.~Piroli, V.~Dallabetta, M.~Walessa, D.~A. Meissner, J.~Kopp, and K.~C.~J.
  Dietmayer, ``Detection of condensed vehicle gas exhaust in lidar point
  clouds,'' \emph{2022 IEEE 25th International Conference on Intelligent
  Transportation Systems (ITSC)}, pp. 600--606, 2022.

\bibitem{hendrycks2016baseline}
D.~Hendrycks and K.~Gimpel, ``A baseline for detecting misclassified and
  out-of-distribution examples in neural networks,'' \emph{arXiv preprint
  arXiv:1610.02136}, 2016.

\bibitem{Piroli2022Robust3O}
A.~Piroli, V.~Dallabetta, M.~Walessa, D.~A. Meissner, J.~Kopp, and K.~C.~J.
  Dietmayer, ``Robust 3d object detection in cold weather conditions,''
  \emph{2022 IEEE Intelligent Vehicles Symposium (IV)}, pp. 287--294, 2022.

\bibitem{du2022vos}
X.~Du, Z.~Wang, M.~Cai, and Y.~Li, ``Vos: Learning what you don’t know by
  virtual outlier synthesis,'' \emph{Proceedings of the International
  Conference on Learning Representations}, 2022.

\bibitem{feng2018towards}
D.~Feng, L.~Rosenbaum, and K.~Dietmayer, ``Towards safe autonomous driving:
  Capture uncertainty in the deep neural network for lidar 3d vehicle
  detection,'' in \emph{2018 21st international conference on intelligent
  transportation systems (ITSC)}.\hskip 1em plus 0.5em minus 0.4em\relax IEEE,
  2018, pp. 3266--3273.

\bibitem{hau2021shadow}
Z.~Hau, S.~Demetriou, L.~Mu{\~n}oz-Gonz{\'a}lez, and E.~C. Lupu,
  ``Shadow-catcher: Looking into shadows to detect ghost objects in autonomous
  vehicle 3d sensing,'' in \emph{Computer Security--ESORICS 2021: 26th European
  Symposium on Research in Computer Security, Darmstadt, Germany, October 4--8,
  2021, Proceedings, Part I 26}.\hskip 1em plus 0.5em minus 0.4em\relax
  Springer, 2021, pp. 691--711.

\bibitem{huang2022out}
C.~Huang, V.~Abdelzad, C.~G. Mannes, L.~Rowe, B.~Therien, R.~Salay,
  K.~Czarnecki, \emph{et~al.}, ``Out-of-distribution detection for lidar-based
  3d object detection,'' in \emph{2022 IEEE 25th International Conference on
  Intelligent Transportation Systems (ITSC)}.\hskip 1em plus 0.5em minus
  0.4em\relax IEEE, 2022, pp. 4265--4271.

\bibitem{geiger2013vision}
A.~Geiger, P.~Lenz, C.~Stiller, and R.~Urtasun, ``Vision meets robotics: The
  kitti dataset,'' \emph{The International Journal of Robotics Research},
  vol.~32, no.~11, pp. 1231--1237, 2013.

\bibitem{sun2020scalability}
P.~Sun, H.~Kretzschmar, X.~Dotiwalla, A.~Chouard, V.~Patnaik, P.~Tsui, J.~Guo,
  Y.~Zhou, Y.~Chai, B.~Caine, \emph{et~al.}, ``Scalability in perception for
  autonomous driving: Waymo open dataset,'' in \emph{Proceedings of the
  IEEE/CVF conference on computer vision and pattern recognition}, 2020, pp.
  2446--2454.

\bibitem{shi2023pv}
S.~Shi, L.~Jiang, J.~Deng, Z.~Wang, C.~Guo, J.~Shi, X.~Wang, and H.~Li,
  ``Pv-rcnn++: Point-voxel feature set abstraction with local vector
  representation for 3d object detection,'' \emph{International Journal of
  Computer Vision}, vol. 131, no.~2, pp. 531--551, 2023.

\bibitem{qi2017pointnet++}
C.~R. Qi, L.~Yi, H.~Su, and L.~J. Guibas, ``Pointnet++: Deep hierarchical
  feature learning on point sets in a metric space,'' \emph{Advances in neural
  information processing systems}, vol.~30, 2017.

\bibitem{zhou2018voxelnet}
Y.~Zhou and O.~Tuzel, ``Voxelnet: End-to-end learning for point cloud based 3d
  object detection,'' in \emph{Proceedings of the IEEE conference on computer
  vision and pattern recognition}, 2018, pp. 4490--4499.

\bibitem{yan2018second}
Y.~Yan, Y.~Mao, and B.~Li, ``Second: Sparsely embedded convolutional
  detection,'' \emph{Sensors}, vol.~18, no.~10, p. 3337, 2018.

\bibitem{shi2020pv}
S.~Shi, C.~Guo, L.~Jiang, Z.~Wang, J.~Shi, X.~Wang, and H.~Li, ``Pv-rcnn:
  Point-voxel feature set abstraction for 3d object detection,'' in
  \emph{Proceedings of the IEEE/CVF Conference on Computer Vision and Pattern
  Recognition}, 2020, pp. 10\,529--10\,538.

\bibitem{zhang2023mixture}
J.~Zhang, N.~Inkawhich, R.~Linderman, Y.~Chen, and H.~Li, ``Mixture outlier
  exposure: Towards out-of-distribution detection in fine-grained
  environments,'' in \emph{Proceedings of the IEEE/CVF Winter Conference on
  Applications of Computer Vision}, 2023, pp. 5531--5540.

\bibitem{zhang2017mixup}
H.~Zhang, M.~Cisse, Y.~N. Dauphin, and D.~Lopez-Paz, ``mixup: Beyond empirical
  risk minimization,'' in \emph{International Conference on Learning
  Representations}, 2018.

\bibitem{yun2019cutmix}
S.~Yun, D.~Han, S.~J. Oh, S.~Chun, J.~Choe, and Y.~Yoo, ``Cutmix:
  Regularization strategy to train strong classifiers with localizable
  features,'' in \emph{Proceedings of the IEEE/CVF international conference on
  computer vision}, 2019, pp. 6023--6032.

\bibitem{bogdoll2022anomaly}
D.~Bogdoll, M.~Nitsche, and J.~M. Z{\"o}llner, ``Anomaly detection in
  autonomous driving: A survey,'' in \emph{Proceedings of the IEEE/CVF
  conference on computer vision and pattern recognition}, 2022, pp. 4488--4499.

\bibitem{lee2018simple}
K.~Lee, K.~Lee, H.~Lee, and J.~Shin, ``A simple unified framework for detecting
  out-of-distribution samples and adversarial attacks,'' \emph{Advances in
  neural information processing systems}, vol.~31, 2018.

\bibitem{dinh2016density}
L.~Dinh, J.~Sohl-Dickstein, and S.~Bengio, ``Density estimation using real
  nvp,'' \emph{arXiv preprint arXiv:1605.08803}, 2016.

\bibitem{openpcdet2020}
O.~D. Team, ``Openpcdet: An open-source toolbox for 3d object detection from
  point clouds,'' \url{https://github.com/open-mmlab/OpenPCDet}, 2020.

\bibitem{kingma2014adam}
D.~P. Kingma and J.~Ba, ``Adam: A method for stochastic optimization,''
  \emph{arXiv preprint arXiv:1412.6980}, 2014.

\end{thebibliography}

\end{document}